\documentclass[conference]{IEEEtran}
\IEEEoverridecommandlockouts
\usepackage{cite}
\usepackage{amsmath,amssymb,amsfonts}
\usepackage{algorithmic}
\usepackage{graphicx}
\usepackage{textcomp}
\usepackage{xcolor}
\usepackage{hyperref}
\def\BibTeX{{\rm B\kern-.05em{\sc i\kern-.025em b}\kern-.08em
    T\kern-.1667em\lower.7ex\hbox{E}\kern-.125emX}}
\begin{document}

\title{Enhancing Airline Customer Satisfaction: A Machine Learning and Causal Analysis Approach}

\author{\IEEEauthorblockN{Tejas Mirthipati}
\IEEEauthorblockA{\textit{Georgia Institute Of Technology} \\
Atlanta, USA \\
tmirthipati3@gatech.edu}
}

\maketitle

\begin{abstract}
This study explores the enhancement of customer satisfaction in the airline industry, a critical factor for retaining customers and building brand reputation, which are vital for revenue growth. Utilizing a combination of machine learning and causal inference methods, we examine the specific impact of service improvements on customer satisfaction, with a focus on the online boarding pass experience. Through detailed data analysis involving several predictive and causal models, we demonstrate that improvements in the digital aspects of customer service significantly elevate overall customer satisfaction. This paper highlights how airlines can strategically leverage these insights to make data-driven decisions that enhance customer experiences and, consequently, their market competitiveness.
\end{abstract}


\section{Introduction}
\subsection{Industry Context}
The COVID-19 pandemic has significantly disrupted the global airline industry, leading to unprecedented revenue losses and slow recovery rates. As airlines endeavor to regain their pre-pandemic financial stability, understanding and addressing customer pain points becomes imperative for maintaining competitive success. This research focuses on analyzing customer satisfaction trends and identifying actionable insights that can enhance service quality across the industry.

\subsection{Problem Statement}
Recent stock performance trends underscore the pervasive challenges within the airline industry. For instance, Delta Air Lines, along with other major carriers such as United Airlines, American Airlines, and Southwest Airlines, has not returned to its pre-2020 stock valuation. This pattern indicates a uniform struggle across the sector, suggesting a widespread need for strategic adjustments in customer service approaches\cite{Atems2021}.

\begin{figure}[htbp]
\centerline{\includegraphics[width=0.8\linewidth]{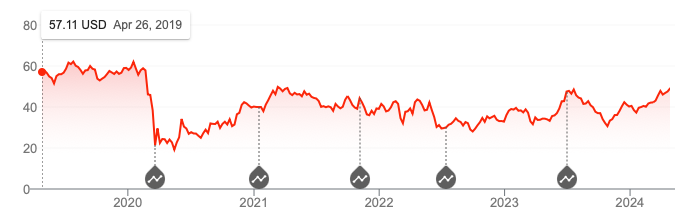}}
\caption{Five-Year Stock Performance of Delta Air Lines (2020-2024).}
\label{fig:delta_stock}
\end{figure}

\subsection{Research Objectives}
This study leverages advanced machine learning and causal inference techniques to examine how specific service improvements, particularly in the online boarding pass process, impact overall customer satisfaction. By integrating quantitative data analysis with model-driven predictions, we aim to provide airlines with evidence-based recommendations for improving customer experiences and, consequently, their operational performance.

\subsection{Service-Profit Chain}
A growing number of companies are placing emphasis on the Service-Profit Chain model, which underscores the importance of treating employees and customers well. This model suggests that enhancing customer satisfaction and loyalty can significantly impact revenue  \cite{HBR2008}.

\begin{figure}[htbp]
\centerline{\includegraphics[width=0.8\linewidth]{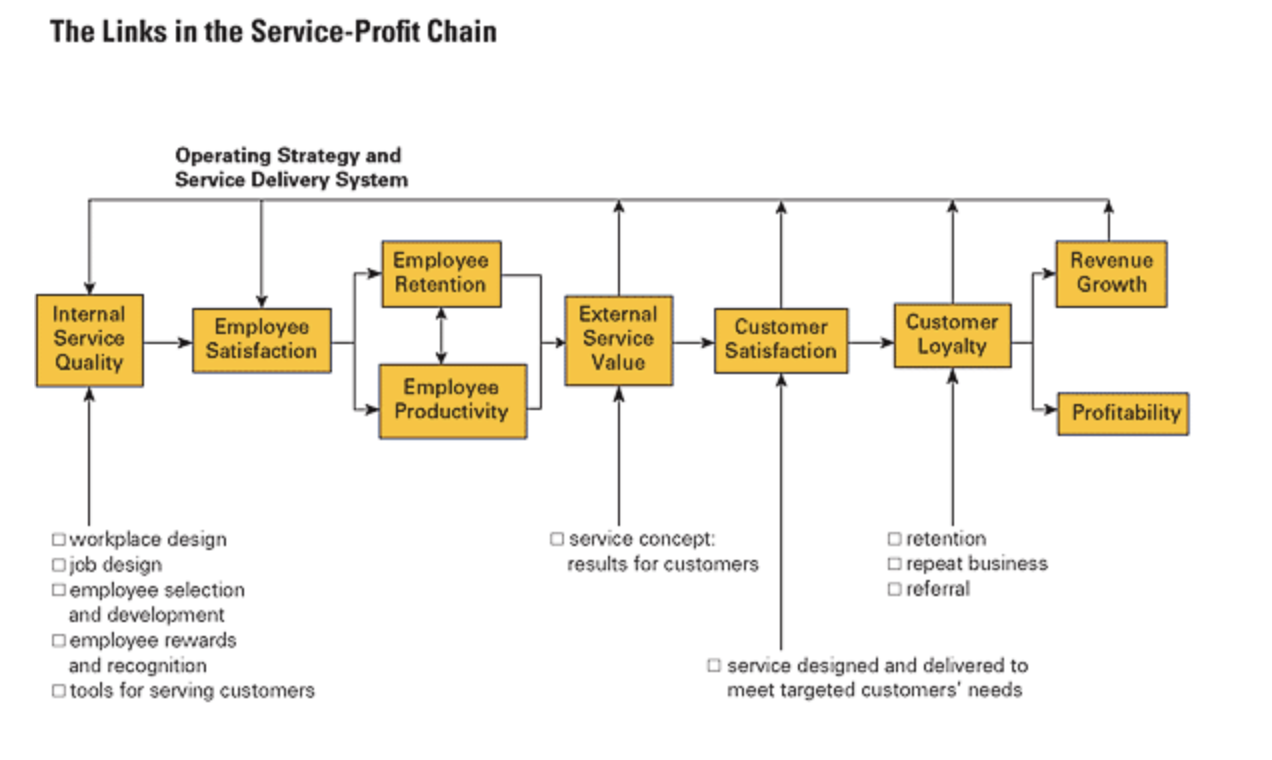}}
\caption{The Service-Profit Chain Model Illustrating the Link between Service Quality and Revenue.}
\label{fig:service_profit_chain}
\end{figure}

\subsection{Research Methodology}
In this project, we focus on the latter half of the Service-Profit Chain, starting with service value, to determine what airlines can improve to directly affect customer satisfaction. We address this through a twofold investigative process:
\begin{enumerate}
    \item How can we leverage machine learning and causal models to help an airline improve customer satisfaction?
    \item What specific features should airlines target to enhance customer satisfaction?
\end{enumerate}

Airlines collect a vast amount of data on customer satisfaction through surveys that gauge various aspects of the flight experience, such as convenience of departure time, leg room, and departure/arrival delays. At the end of these surveys, customers are also asked to rate their overall satisfaction with their experience. We plan to use such survey data to identify the best predictors of overall satisfaction by training classification models with a variety of methods covered in class. To further this analysis, we will introduce a Python library to perform a causal analysis, as it is crucial for airlines to understand where they can make improvements to effectively increase overall customer satisfaction.

\section{Data Collection and Preprocessing}

\subsection{Data Background}
The dataset titled “Airline Passenger Satisfaction,” sourced from Kaggle, serves as the foundation for this study. It comprises over 100,000 observations, approximately 15 MB in size, and contains 22 predictors of customer satisfaction, including both continuous and categorical variables. The response variable is binary, where `1` indicates a satisfied customer and `0` indicates a neutral or dissatisfied customer \cite{kaggleDataset}. A snapshot of the raw data is presented below.

\begin{figure}[htbp]
\centerline{\includegraphics[width=0.8\linewidth]{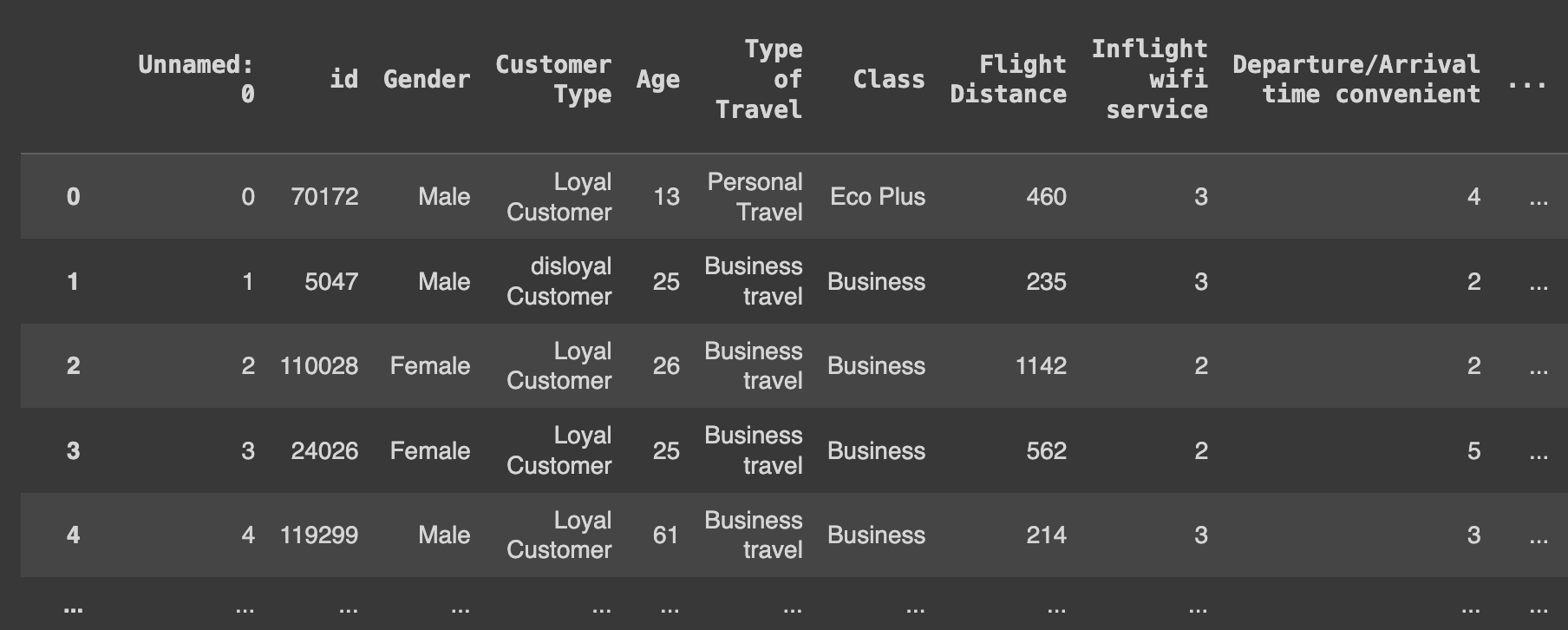}}
\caption{Snapshot of Raw Data}
\label{fig:raw_data_snapshot}
\end{figure}

\subsection{Data Preprocessing}
The dataset was initially examined for balance between the satisfied and neutral/dissatisfied customer responses. Given the fairly even distribution, accuracy score was deemed an appropriate metric for model evaluation. The preprocessing steps involved removing duplicate entries and unnecessary columns such as "unnamed" and "id." Null values were identified and imputed with the median to address data skewness, as shown in the box plot for the arrival/departure delay data below.

\begin{figure}[htbp]
\centerline{\includegraphics[width=0.6\linewidth]{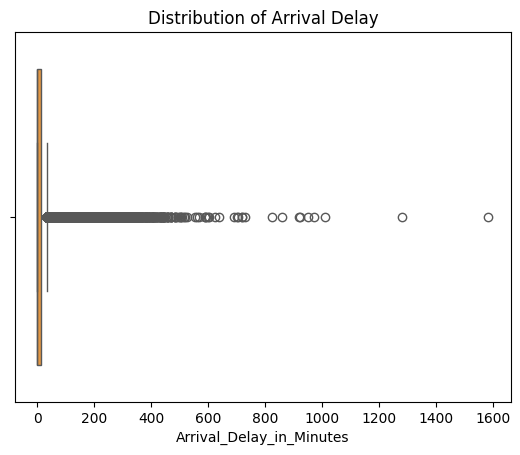}}
\caption{Box Plot of Arrival/Departure Delays Highlighting Outliers}
\label{fig:arrival_departure_delays}
\end{figure}

Ordinal categorical data from the customer survey were label encoded to maintain the natural ordering of the ratings from zero to five. For numerical data normalization, the Min Max scaler was employed to scale the data within the range [0,1], thus preserving the distribution of outlier values which were considered significant to the analysis.

Concerns of multicollinearity were addressed by examining scatter plots of potentially collinear variables. An example scatter plot demonstrating a strong linear relationship between two predictors is shown below; based on this analysis, one of the variables was excluded from the logistic regression model. Tree-based models, such as random forests and decision trees, were not adjusted for multicollinearity due to their inherent robustness to such issues.

\begin{figure}[htbp]
\centerline{\includegraphics[width=0.9\linewidth]{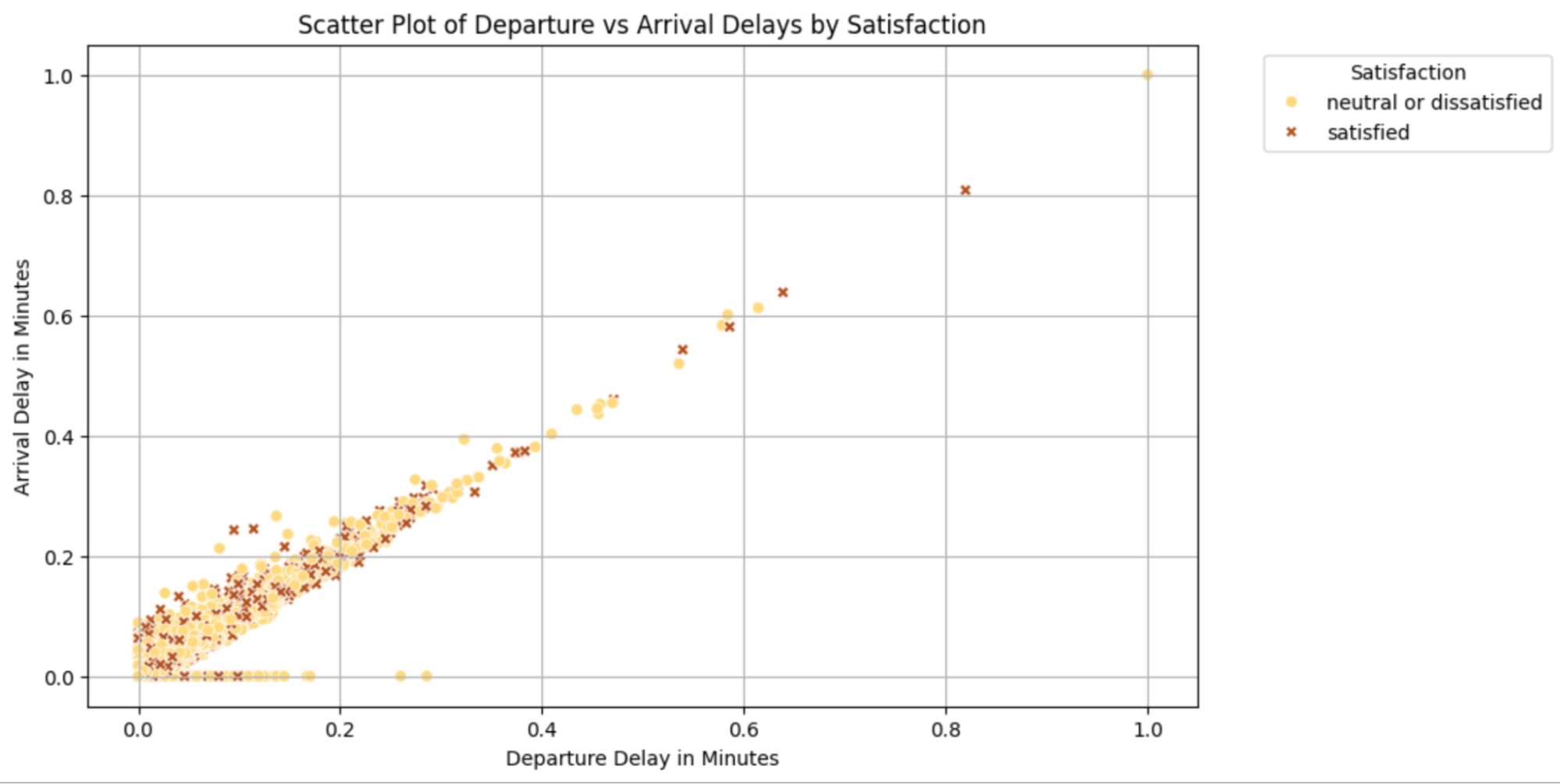}}
\caption{Scatter Plot Illustrating Collinearity Between Two Predictors}
\label{fig:collinearity_scatter_plot}
\end{figure}

Finally, a snapshot of the processed data, ready for further analysis, is depicted below.

\begin{figure}[htbp]
\centerline{\includegraphics[width=0.8\linewidth]{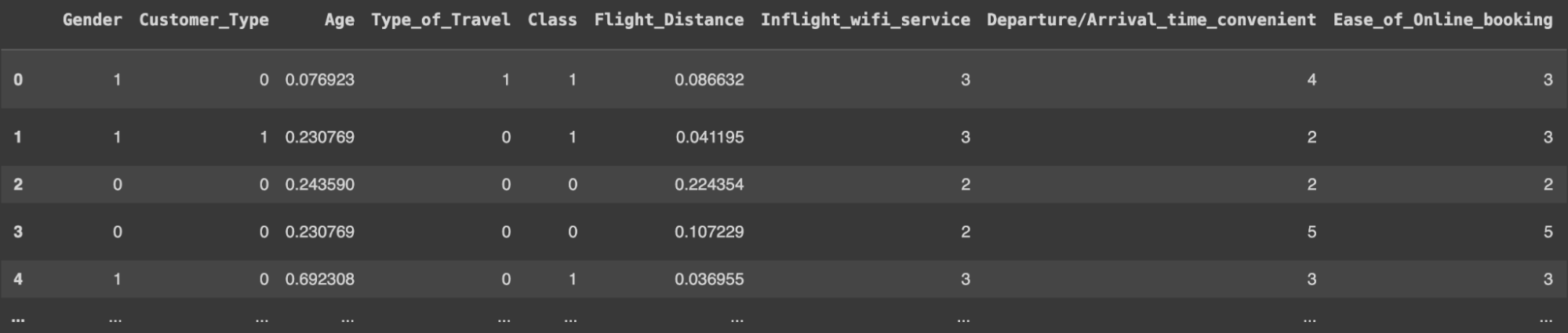}}
\caption{Snapshot of Processed Data}
\label{fig:processed_data_snapshot}
\end{figure}

\subsection{Exploratory Data Analysis}
Exploratory Data Analysis (EDA) provided several key insights crucial for understanding the factors influencing customer satisfaction before advancing to predictive modeling. Our analysis encompassed various demographic and service-related factors.

\subsubsection{Age Distribution Analysis}
\begin{figure}[htbp]
\centerline{\includegraphics[width=0.75\linewidth]{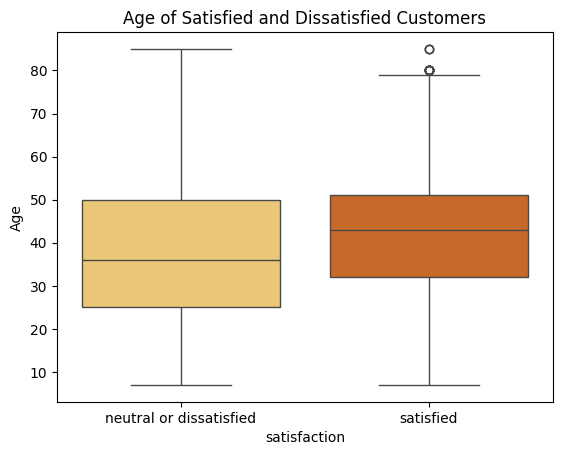}}
\caption{Age Distribution of Airline Customers by Satisfaction Level}
\label{fig:age_distribution}
\end{figure}

Figure \ref{fig:age_distribution} presents a box and whisker plot comparing the age distribution between neutral/dissatisfied customers and satisfied customers. It reveals that the median age of satisfied customers is approximately 10 years older than that of dissatisfied customers, indicating that age may play a significant role in satisfaction. Additionally, there is a greater variance in the ages of dissatisfied customers.

\subsubsection{Customer Satisfaction by Travel Class}
\begin{figure}[htbp]
\centerline{\includegraphics[width=0.75\linewidth]{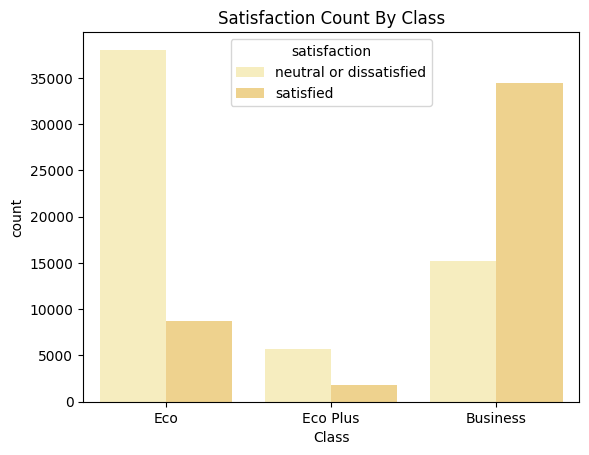}}
\caption{Customer Satisfaction by Class of Travel}
\label{fig:class_satisfaction}
\end{figure}

Analysis of customer satisfaction by class of travel (Figure \ref{fig:class_satisfaction}) shows that a higher proportion of dissatisfied customers travel in Economy class, whereas Business class travelers report higher satisfaction. This suggests that service level and amenities associated with higher travel classes contribute to increased satisfaction.

\subsubsection{Flight Delay Tolerance by Flight Length}
\begin{figure}[htbp]
\centerline{\includegraphics[width=0.75\linewidth]{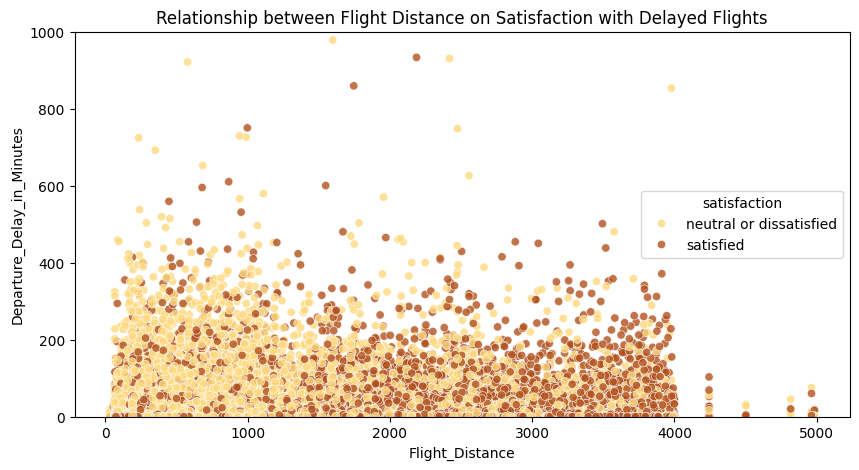}}
\caption{Impact of Departure Delays on Customer Satisfaction by Flight Distance}
\label{fig:delay_distance}
\end{figure}

A scatter plot of departure delay against flight distance (Figure \ref{fig:delay_distance}) illustrates that customers on longer flights exhibit greater tolerance for departure delays, with a higher satisfaction rate among these passengers despite increased delays.

\subsubsection{Loyalty Membership and Satisfaction}
\begin{figure}[htbp]
\centerline{\includegraphics[width=0.75\linewidth]{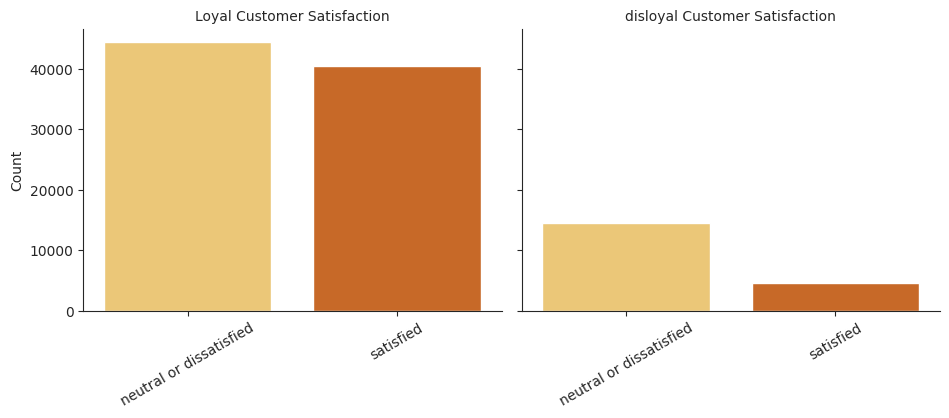}}
\caption{Customer Satisfaction Among Loyalty and Non-loyalty Members}
\label{fig:loyalty_satisfaction}
\end{figure}

Our analysis indicates that loyalty members tend to show a more balanced satisfaction rate compared to non-members, as illustrated in Figure \ref{fig:loyalty_satisfaction}. This highlights the effectiveness of loyalty programs in mitigating dissatisfaction.

\subsubsection{Online Boarding and Departure/Arrival Time Satisfaction}
\begin{figure}[htbp]
\centerline{\includegraphics[width=0.75\linewidth]{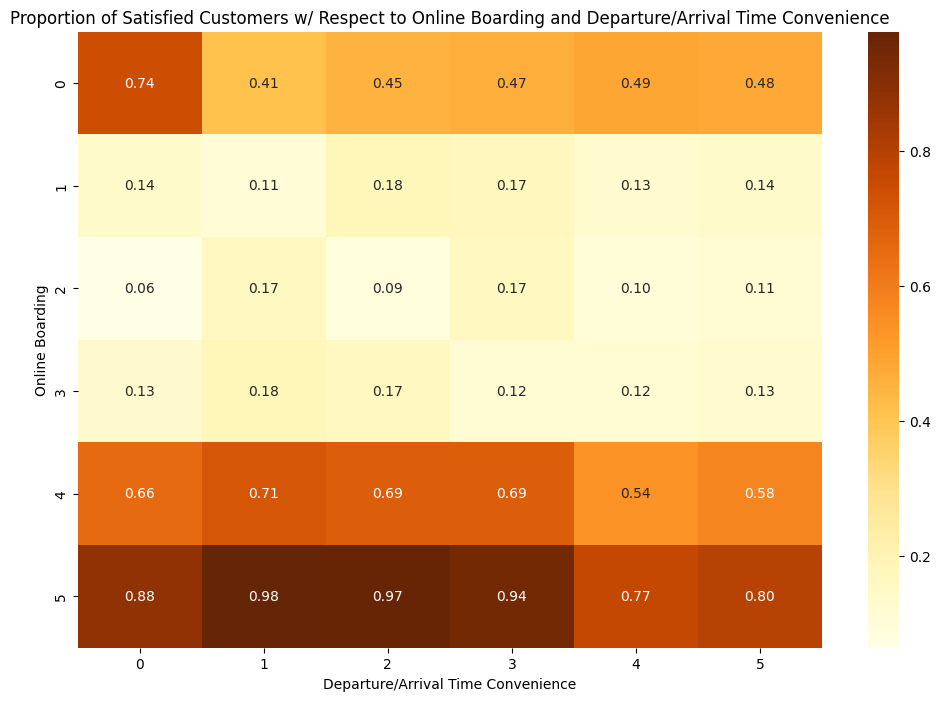}}
\caption{Correlation Heat Map of Online Boarding and Departure/Arrival Time Satisfaction}
\label{fig:online_boarding_heatmap}
\end{figure}

Finally, the heat map in Figure \ref{fig:online_boarding_heatmap} shows a strong positive correlation between online boarding satisfaction and overall customer satisfaction. In contrast, satisfaction with departure/arrival times shows a less pronounced impact.

\section{Model Description and Training}
This section details the application of various machine learning models to understand the predictors of customer satisfaction in airlines, followed by a causal analysis to examine the effect of specific features on satisfaction levels.

\subsection{Classification Models}
We employed six different machine learning models for binary classification of customer satisfaction: Decision Trees, Random Forests, Gradient Boosting, K-Nearest Neighbors (KNN), Support Vector Machines (SVM), and Logistic Regression. Each model was specifically configured and optimized through a rigorous training process.

\subsubsection{Training Process}
The models were initially trained using an 80-20 split of the data, where 80\% was allocated for training and the remaining 20\% served as a holdout test set. We utilized GridSearchCV with 5-fold cross-validation to fine-tune hyperparameters, aiming to balance the models' complexity and generalization capabilities.

\subsubsection{Parameter Sensitivity Analysis}
Parameter sensitivity curves were generated to visually assess how changes in key parameters impact model performance, aiding in the selection of the optimal model configurations.

\paragraph{Model Specifics}
Each model was set up with distinct configurations and parameter ranges as follows:
\begin{itemize}
    \item Decision Tree: Max Depth = 14, Parameter Range Tested: 1 to 20
    \item Random Forest: Max Depth = 25, Parameter Range Tested: 10 to 25
    \item Gradient Boosting: Max Depth = 9, Parameter Range Tested: 3 to 11
    \item K-Nearest Neighbors (KNN): n\_neighbors = 5, Parameter Range Tested: 1 to 10
    \item Support Vector Machine (SVM): C = 100, Parameter Range Tested: 0.001 to 100
    \item Logistic Regression: C = 207, Parameter Range Tested: 0.0001 to 10,000 (logarithmic scale)
\end{itemize}

\section{Causal Analysis Using the CausalLib Package}
The objective of our causal analysis is to understand the true impact of service improvements, specifically the satisfaction with online boarding passes, on overall customer satisfaction. Unlike machine learning models that primarily predict outcomes based on feature importance, our causal analysis seeks to establish a definitive cause-and-effect relationship \cite{Pearl2009,Rubin2005}.

\subsection{Defining the Treatment}
Using the CausalLib package, which is designed for analyzing observational data, we model the causal effect of being satisfied with the online boarding process. Satisfaction is quantified through survey ratings, with scores of 4 or 5 on a 1-5 scale indicating satisfaction (treatment group) and scores of 1 to 3 indicating dissatisfaction (control group). These scores are used to create a binary treatment variable critical for our causal inference \cite{Hernan2020}.

\subsection{Propensity Score Estimation}
The initial step in our causal modeling involves calculating the propensity scores, which estimate the likelihood of receiving the treatment based on observed covariates. This involves using robust machine learning models capable of handling classification tasks effectively, such as Logistic Regression or Random Forest \cite{Rosenbaum1983,Breiman2001}.

\subsection{Weight Calculation and Model Adjustment}
\subsubsection{Inverse Propensity Weighting}
Weights are calculated as the inverse of the propensity scores to balance the distribution of covariates across the treatment and control groups:
\begin{equation}
\text{weight} = \frac{1}{\text{propensity score}}
\end{equation}
This approach adjusts for potential biases due to non-random treatment assignment, aligning with standard practices in causal inference to ensure robust effect estimation \cite{Hirano2001}.

\subsubsection{Handling Weight Extremes}
To manage extreme weights, which can skew results, we explored using parameters like \texttt{clip\_min} and \texttt{clip\_max} to cap weights at certain thresholds. However, our analysis showed that excluding these clippings provided a more conservative estimate of the causal effect, reinforcing the robustness of our findings without the need for such adjustments  \cite{Austin2015}.

\subsection{Horvitz-Thompson Estimator}
Using the Horvitz-Thompson estimator, we compute the average treatment effect by weighting the observed outcomes by their inverse propensity weights:
\begin{equation}
\hat{E}[Y_a] = \frac{\sum_{i: A_i=a} w_i y_i}{\sum_{i: A_i=a} w_i}
\end{equation}
where \(A_i\) indicates the treatment assignment, \(w_i\) denotes the weights, and \(y_i\) is the outcome of interest \cite{Horvitz1952}.

\subsection{Effect Measurement}
The treatment effect is evaluated by comparing the weighted average outcomes between the treatment and control groups. This analysis provides a clear measure of the impact that satisfaction with online boarding passes has on overall customer satisfaction, offering valuable insights for airline service improvement strategies \cite{Imbens2009}.

\section{Experimental Results}

\subsection{Evaluation Metrics and Model Performance}
The primary metric for evaluating model performance in this study was accuracy, selected due to:
\begin{itemize}
    \item \textbf{Balanced Classes:} The target variable, customer satisfaction, exhibited balanced classes, allowing accuracy to effectively reflect model performance without bias.
    \item \textbf{Simplicity and Relevance:} Since specific costs of misclassification (false negatives and false positives) were not critical for this project, simpler metrics like accuracy were deemed sufficient over precision and recall.
\end{itemize}

The performance of various classification models is summarized in the table below, displaying both cross-validation and test accuracies:

\begin{table}[htbp]
\centering
\begin{tabular}{|l|c|c|}
\hline
\textbf{Model} & \textbf{Cross-Validation Accuracy} & \textbf{Test Accuracy} \\
\hline
SVM & 0.96 & 0.95 \\
Decision Tree & 0.95 & 0.92 \\
Random Forest & 0.96 & 0.95 \\
KNN & 0.93 & 0.92 \\
Gradient Boosting & 0.96 & 0.95 \\
Logistic Regression & 0.87 & 0.87 \\
\hline
\end{tabular}
\caption{Model Performance Summary}
\label{tab:model_performance}
\end{table}

\subsection{Learning Curves}
Analysis of learning curves for each model revealed:
\begin{itemize}
    \item \textbf{Training Accuracy:} Generally starts high and may decrease slightly as more training data is added.
    \item \textbf{Validation Accuracy:} Typically begins lower but increases with additional training data, suggesting improved model generalization.
\end{itemize}

\subsection{Model Selection and Interpretability}
Despite high accuracy scores from SVM, Random Forest, and Gradient Boosting, a Decision Tree was selected as the primary model due to its interpretability. This feature is crucial for understanding the factors influencing customer satisfaction and aligns with the project's goal to provide actionable insights \cite{Breiman1984}.

\subsubsection{Feature Importance}
The Decision Tree's feature importance, calculated using Gini Impurity, indicated that 'Online Boarding' and 'In-Flight Wifi Service' were significant predictors of satisfaction. This importance is visualized in the bar chart below \cite{Lundberg2017}:

% Assuming the chart is included here as an image
\begin{figure}[htbp]
\centerline{\includegraphics[width=0.75\linewidth]{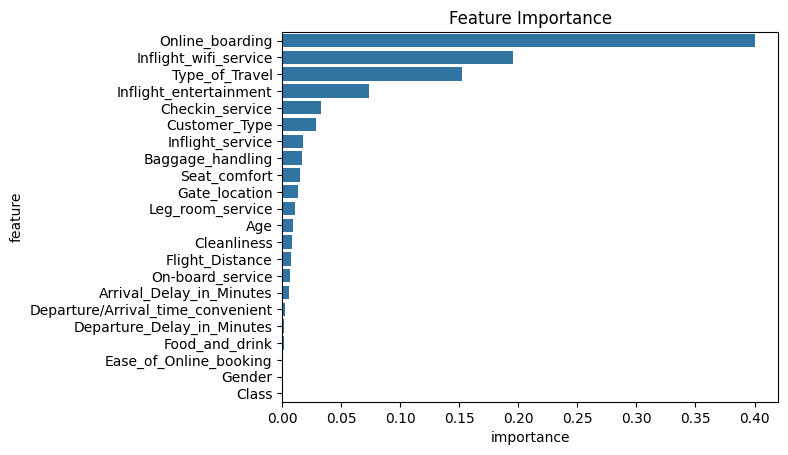}}
\caption{Feature Importance in Decision Tree}
\label{fig:feature_importance}
\end{figure}

\subsubsection{SHAP Summary Plot}
The SHAP summary plot provides deeper insights into how individual feature values affect the prediction outcome. For example, higher ratings for 'Online Boarding' significantly increase the likelihood of satisfaction, as depicted by the positive SHAP values in the plot below \cite{Lundberg2017}:

% Assuming the SHAP plot is included here as an image
\begin{figure}[htbp]
\centerline{\includegraphics[width=0.75\linewidth]{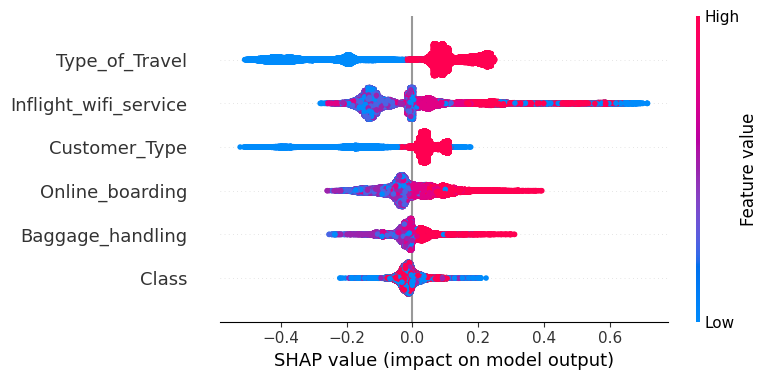}}
\caption{SHAP Summary Plot for Online Boarding}
\label{fig:shap_summary}
\end{figure}

\subsection{Causal Model: Weight Balance Analysis and Model Performance}
To assess the effectiveness of the weighting process used in our causal models, we examined the covariate balance achieved by Logistic Regression, Random Forest, and XGBoost through love graphs  \cite{Austin2011}. These graphs visually compare the absolute standard mean differences between treatment groups by covariate, both before and after weighting.

\begin{figure}[htbp]
\centerline{\includegraphics[width=0.6\linewidth]{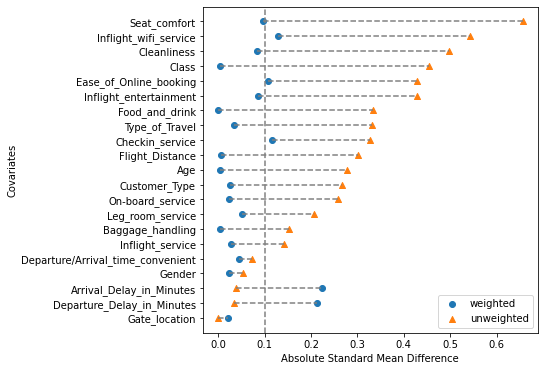}}
\caption{Logistic Regression Model Love Graph showing improved covariate balance after weighting.}
\label{fig:logistic_regression_love_graph}
\end{figure}

\begin{figure}[htbp]
\centerline{\includegraphics[width=0.6\linewidth]{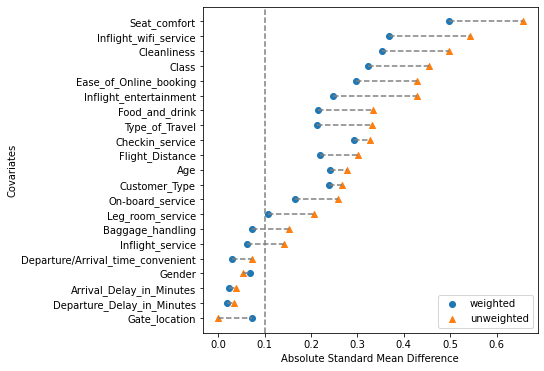}}
\caption{Random Forest Classifier Love Graph indicating adjustments in covariate balance.}
\label{fig:random_forest_love_graph}
\end{figure}

\begin{figure}[htbp]
\centerline{\includegraphics[width=0.6\linewidth]{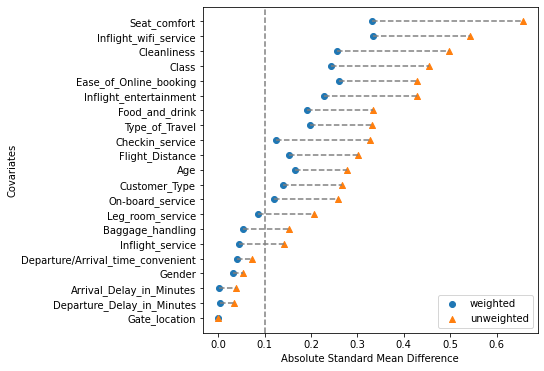}}
\caption{XGBoost Love Graph depicting balance effectiveness.}
\label{fig:xgboost_love_graph}
\end{figure}

The Logistic Regression model exhibited the most significant improvement in balancing covariates, with the smallest mean differences, suggesting it is the most effective model for adjusting confounders in our analysis.

\begin{figure}[htbp]
\centerline{\includegraphics[width=0.6\linewidth]{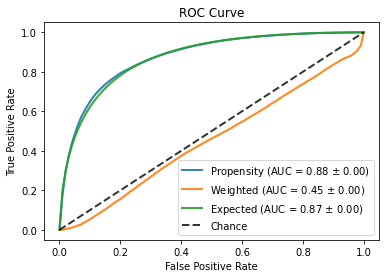}}
\caption{ROC Curve for Logistic Regression illustrating effective propensity score modeling.}
\label{fig:logistic_regression_roc}
\end{figure}

\begin{figure}[htbp]
\centerline{\includegraphics[width=0.6\linewidth]{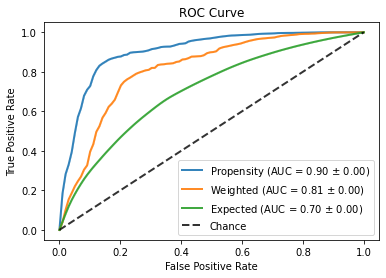}}
\caption{ROC Curve for Random Forest, showing higher predictive accuracy.}
\label{fig:random_forest_roc}
\end{figure}

\begin{figure}[htbp]
\centerline{\includegraphics[width=0.6\linewidth]{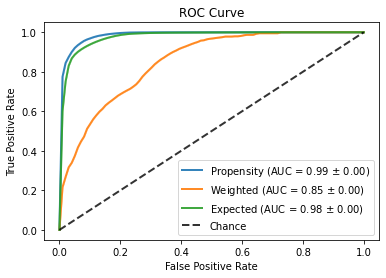}}
\caption{ROC Curve for XGBoost indicating potential overfitting.}
\label{fig:xgboost_roc}
\end{figure}

\paragraph{ROC Curve Analysis}
The Receiver Operating Characteristic (ROC) curves for each model were analyzed to further evaluate the predictive accuracy and the propensity score model’s effectiveness, as shown in Figures \ref{fig:logistic_regression_roc}, \ref{fig:random_forest_roc}, and \ref{fig:xgboost_roc}. This approach is crucial to ensure the model accurately estimates the probability of treatment assignment \cite{Fawcett2006}.

The curves show that while Random Forest and XGBoost provide higher predictive accuracy, Logistic Regression achieves a balance closer to the ideal chance line, indicating a more accurate estimation of the treatment effect.

\subsection{Results and Interpretation of Treatment Effects}
The treatment effect was evaluated by comparing weighted average outcomes between the treatment and control groups, utilizing methods well-established in the literature for their reliability in causal effect estimation \cite{Imbens2015}.

\subsubsection{Effect Size Estimation}
The estimated causal effect of satisfaction with online boarding on customer satisfaction was 0.236, suggesting a substantial potential for improving overall satisfaction by enhancing the online boarding experience \cite{Angrist2008}.

\subsubsection{Comparative Analysis with In-Flight Wifi}
Similar analysis for in-flight wifi satisfaction resulted in an estimated causal effect of 0.139, indicating it as another significant but less impactful area compared to online boarding \cite{Ho2007}.

\paragraph{Marginal Balance Outcome}
The analysis also explored the marginal balance outcome, which provides an estimate of the treatment effect without the weighting adjustments, showing the importance of using robust weighting methods to obtain unbiased causal estimates \cite{Morgan2012}.

\begin{equation}
\text{Marginal Effect for Online Boarding} = 0.574
\end{equation}

\begin{equation}
\text{Marginal Effect for In-Flight Wifi} = 0.443
\end{equation}

These unadjusted effects were significantly higher than those adjusted by IPW, underscoring the importance of weighting in obtaining unbiased causal estimates.

\section{Conclusion}
This project employed a robust two-step approach combining machine learning and causal analysis to explore the determinants of customer satisfaction within the airline industry. Our comprehensive analysis has yielded significant insights that can directly inform strategies to enhance airline services and, consequently, improve brand reputation.

\subsection{Key Findings}
We utilized a variety of advanced classification models, such as decision trees, random forests, gradient boosting, and logistic regression, to identify critical predictors of customer satisfaction. Our findings highlighted that satisfaction with online boarding passes and in-flight wifi services are crucial drivers of overall customer contentment. These features were consistently significant across different models, underscoring their importance in the customer service experience.

\subsection{Causal Analysis}
Further, the application of the CausalLib package allowed us to extend beyond predictive analytics into causal inference, utilizing observational data to validate the influence of identified predictors. This analysis not only confirmed the importance of these service aspects but also demonstrated a direct causal relationship between improved service features—particularly online boarding—and increased customer satisfaction. Such insights validate the predictive models and provide a stronger basis for recommending specific service enhancements.

\subsection{Recommendations for the Airline Industry}
Based on our findings, we recommend several actionable strategies for airlines aiming to differentiate their services and enhance customer satisfaction:
\begin{itemize}
    \item \textbf{Streamlining Online Boarding:} Simplify the online boarding process to allow passengers to obtain their boarding passes easily without the necessity of downloading an app. This could reduce friction in the travel experience and enhance customer satisfaction.
    \item \textbf{Complimentary In-Flight Wifi:} Offering wifi onboard at no extra charge could significantly enhance the in-flight experience, promoting customer loyalty and potentially increasing the likelihood of repeat business.
\end{itemize}

These strategies are particularly pertinent as airlines continue to navigate the competitive and rapidly changing landscape of the travel industry. Implementing these recommendations could not only improve customer satisfaction but also position airlines as leaders in customer-centric service.

\subsection{Conclusion}
In conclusion, this project illustrates the power of integrating machine learning and causal analysis to derive actionable insights from complex data landscapes. For the airline industry, adopting such data-driven approaches in strategic decision-making can lead to substantial improvements in customer satisfaction and business success. Our study offers a clear roadmap for leveraging technological advancements to enhance service offerings and compete effectively in today's economy.

\end{document}